%% file: main.tex
\documentclass[10pt,twocolumn,letterpaper]{article}

\usepackage{cvpr}              % To produce the CAMERA-READY version
\usepackage{booktabs}
\usepackage{subcaption}
\usepackage{multirow}
\usepackage{makecell}
\usepackage{pifont}
\usepackage{wrapfig}

\input{preamble}

\definecolor{cvprblue}{rgb}{0.21,0.49,0.74} 
\usepackage[pagebackref,breaklinks,colorlinks,citecolor=cvprblue]{hyperref}

\def\paperID{13079} % *** Enter the Paper ID here
\def\confName{CVPR}
\def\confYear{2024}

\newcommand{\ourmethod}{Cohere3D}

\title{\ourmethod: Exploiting Temporal Coherence for Unsupervised Representation Learning of Vision-based Autonomous Driving}

\author{Yichen Xie$^1$\thanks{This work was done when Yichen was an intern at Cruise LLC.}\;, Hongge Chen$^2$, Gregory P. Meyer$^2$, Yong Jae Lee$^2$, Eric M. Wolff$^2$,\\ Masayoshi Tomizuka$^1$, Wei Zhan$^1$, Yuning Chai$^2$, Xin Huang$^2$\\
$^{1}$University of California, Berkeley\quad $^{2}$Cruise LLC\\
}

\begin{document}
\maketitle
\input{sec_arxiv/0_abstract}

\input{sec_arxiv/1_intro}
\input{sec_arxiv/2_related_works}

\input{sec_arxiv/3_methods}
\input{sec_arxiv/4_experiments}
\input{sec_arxiv/5_conclusion}

{
    \small
    \bibliographystyle{ieeenat_fullname}
    \bibliography{main}
}

% WARNING: do not forget to delete the supplementary pages from your submission 
\input{sec_arxiv/X_suppl}

% {
%     \small
%     \bibliographystyle{ieeenat_fullname}
%     \bibliography{main}
% }

\end{document}

%% file: preamble.tex
\usepackage[dvipsnames]{xcolor}

%% file: sec_arxiv/0_abstract.tex
\begin{abstract}
Due to the lack of depth cues in images, multi-frame inputs are important for the success of vision-based perception, prediction, and planning in autonomous driving. Observations from different angles enable the recovery of 3D object states from 2D image inputs if we can identify the same instance in different input frames. However, the dynamic nature of autonomous driving scenes leads to significant changes in the appearance and shape of each instance captured by the camera at different time steps. To this end, we propose a novel contrastive learning algorithm, \ourmethod, to learn coherent instance representations in a long-term input sequence robust to the change in distance and perspective. The learned representation aids in instance-level correspondence across multiple input frames in downstream tasks. In the pretraining stage, the raw point clouds from LiDAR sensors are utilized to construct the long-term temporal correspondence for each instance, which serves as guidance for the extraction of instance-level representation from the vision-based bird's eye-view (BEV) feature map. \ourmethod{} encourages a consistent representation for the same instance at different frames but distinguishes between representations of different instances. We evaluate our algorithm by finetuning the pretrained model on various downstream perception, prediction, and planning tasks. Results show a notable improvement in both data efficiency and task performance.
\end{abstract}

%% file: sec_arxiv/1_intro.tex
\section{Introduction}
\label{sec:intro}

% \begin{itemize}
%     \item The emphasis of vision-based perception and prediction on temporal information
%     \item The key spatio-temporal invariance in the representation learning
%     \item Briefly introduce our method
% \end{itemize}

Recent work in vision-based autonomous driving has demonstrated promising performance in 3D detection~\cite{huang2021bevdet,li2023bevdepth}, map reconstruction~\cite{saha2022translating,liao2022maptr}, prediction~\cite{gu2023vip3d,jiang2022perceive}, and planning~\cite{hu2023planning,jiang2023vad}.
Despite the rich semantics provided by images, the absence of depth information poses a critical challenge in recovering 3D geometries from a single-frame camera input. Multi-frame inputs aggregate the observation of the same instances (or regions) from different perspectives at different time steps through correspondence matching to determine their 3D states. 
% However, due the dynamic nature of the ego-vehicle and surrounding scenes, the relative position of each instance \textit{w.r.t.} the ego-car is likely to vary frame by frame, which makes the same instance exhibit different shapes and appearance at multi-frame image inputs. 
However, the dynamic nature of autonomous driving scenes complicates the extraction of reliable instance-level information across temporal frames. The varying relative positions of objects to the cameras lead to significant alternations in their appearance within the images, presenting a big hurdle in maintaining consistent instance identification over time.
This inconsistency complicates finding an instance-level correspondence across multiple temporal frames, which serves as a fundamental barrier to effective fusion of multi-frame inputs. Thus, \textit{coherent instance-wise representations that are robust to the change of observation viewpoint and distance} are critical for effective temporal fusion.

\begin{figure}[tb]
    \centering
    \includegraphics[width=\linewidth, trim={1cm 10cm 1cm 0},clip]{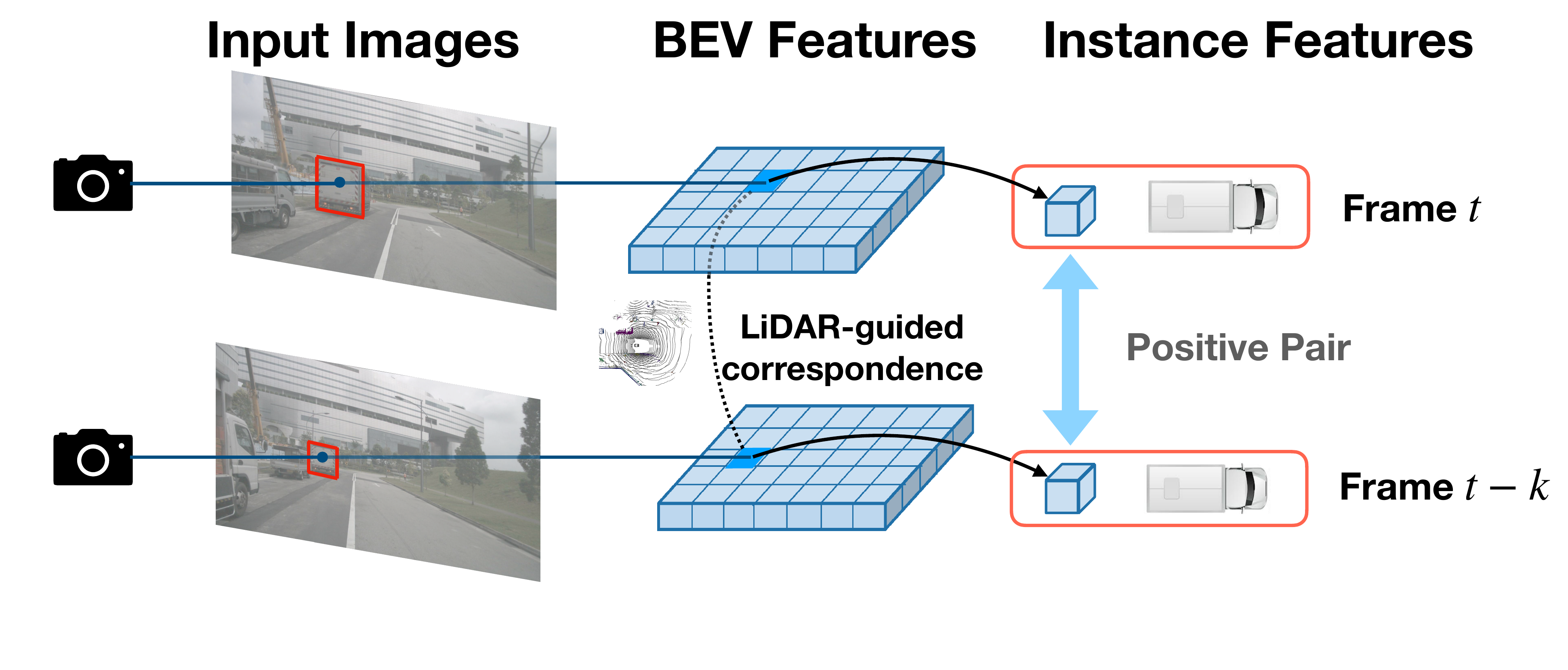}
    \caption{Temporal coherence of instance representations. Our algorithm encourages instance representations in the BEV space to be coherent across time and viewpoints.}
    \label{fig:traj}
    \vspace{-10pt}
\end{figure}

Unfortunately, fine-grained annotations are needed to learn this robust representation, since all the objects and their temporal correspondences have to be identified. Given the immense human effort required for labeling, only a small portion of the sensor data collected by self-driving cars can be annotated. To this end, unsupervised representation learning provides a practical solution for exploiting all of the unlabeled data. 
% In the area of autonomous driving, prior work applies both contrastive learning~\cite{yin2022proposalcontrast,nunes2023cvpr} or masked autoencoder~\cite{tian2023geomae,lin2022bev} strategies, but most of them concentrate on the LiDAR sensor inputs. As the only camera-based approach, Occ-BEV~\cite{min2023occ} ignores the semantics inside a scene\, which limits their performance in combination with state-of-the-art vision-based detectors~\cite{park2022time} (Sec.~\ref{sec:detection}).

For this reason, we propose a novel contrastive learning method~\emph{\ourmethod{}} designed to produce long-term coherent instance representations (Fig.~\ref{fig:traj}) for vision-based perception, prediction, and planning tasks. Our approach leverages the richness of visual data while circumventing the constraints posed by the lack of depth information. Utilizing commonly available raw LiDAR point clouds in the 3D space, our method constructs a long-term temporal correspondence for each instance. This correspondence serves as a foundation for extracting instance-level representations from vision-based bird's eye-view (BEV) feature maps. It thus enhances the model's ability to maintain instance representation consistency across frames.

Our method focuses on the extraction and learning of long-term temporal correspondences, a critical aspect, but often overlooked in previous representation learning methods for autonomous driving tasks. By aligning representations of the same instance across multiple frames and differentiating between distinct instances, our framework achieves a significant improvement in both data efficiency and overall performance in downstream vision-based tasks such as 3D object detection, map reconstruction, motion prediction, and end-to-end planning.

The main contributions of our paper are:
\begin{enumerate}
    \item We identify \textit{long-term instance representation coherence} as a key component for successful vision-based 3D representation learning, which was not well-studied by prior works. 
    \item We introduce a simple yet effective contrastive learning framework, \ourmethod{}, tailored for vision-based autonomous driving tasks, emphasizing the robust learning of long-term temporal correspondences.
    \item We demonstrate superior performance and data efficiency of our pretraining algorithm across vision-based 3D perception, prediction, and planning tasks.
\end{enumerate}

%% file: sec_arxiv/2_related_works.tex
\section{Related Works}
\label{sec:related_works}
\paragraph{Camera-based Autonomous Driving} Cameras are cheap and ubiquitous sensors that are often used in autonomous driving. 
Due to the limited field-of-view of a single camera, the surrounding environment is often captured by multiple cameras from different perspective views. To combine multi-view information in 3D perception tasks, prior works~\cite{huang2021bevdet,li2023bevdepth,liu2023bevfusion} apply an LSS-style~\cite{philion2020lift} projection strategy to transform image features from different perspective views into a unified bird's eye-view space, while others~\cite{wang2022detr3d,liu2022petr,xie2023sparsefusion} utilize a set of 3D queries to learn the inter-view correspondence with transformer-based networks~\cite{carion2020end}. However, without overlapping camera fields-of-views in autonomous driving scenarios, single-frame camera inputs lack reliable depth cues. To this end, a few camera-based perception algorithms~\cite{huang2022bevdet4d,li2022bevformer} exploit several neighboring frames to recover the 3D geometry. SOLOFusion~\cite{park2022time} takes a step further by considering long-term temporal information, which allows camera-based methods~\cite{wang2023exploring,lin2023sparse4d} to achieve performance competitive to LiDAR-based methods~\cite{lang2019pointpillars}. 
% Some work~\cite{liu2023bevfusion,xie2023sparsefusion} also considers cameras as a complement for LiDAR in multi-sensor perception. 
Beyond perception tasks, recent work also develops end-to-end pipelines based on camera inputs. ViP3D~\cite{gu2023vip3d} and PIP~\cite{jiang2022perceive} directly predict motion trajectories using visual information. 
UniAD~\cite{hu2023planning} and VAD~\cite{jiang2023vad} apply a joint training strategy to assist end-to-end planning.

High-quality 3D annotations are essential for all the above algorithms. Despite recent efforts on auto-labeling~\cite{qi2021offboard,yang2021auto4d} or sim-to-real transfer~\cite{chen2020learning} to provide meaningful labels, the offboard perception system is expensive to build and the domain transfer gap still remains an open challenge. Orthogonal to these works, we propose a novel unsupervised representation learning algorithm that makes full use of the abundant unlabeled sensor data.

\begin{figure*}[t]
    \centering
    \includegraphics[width=0.9\linewidth]{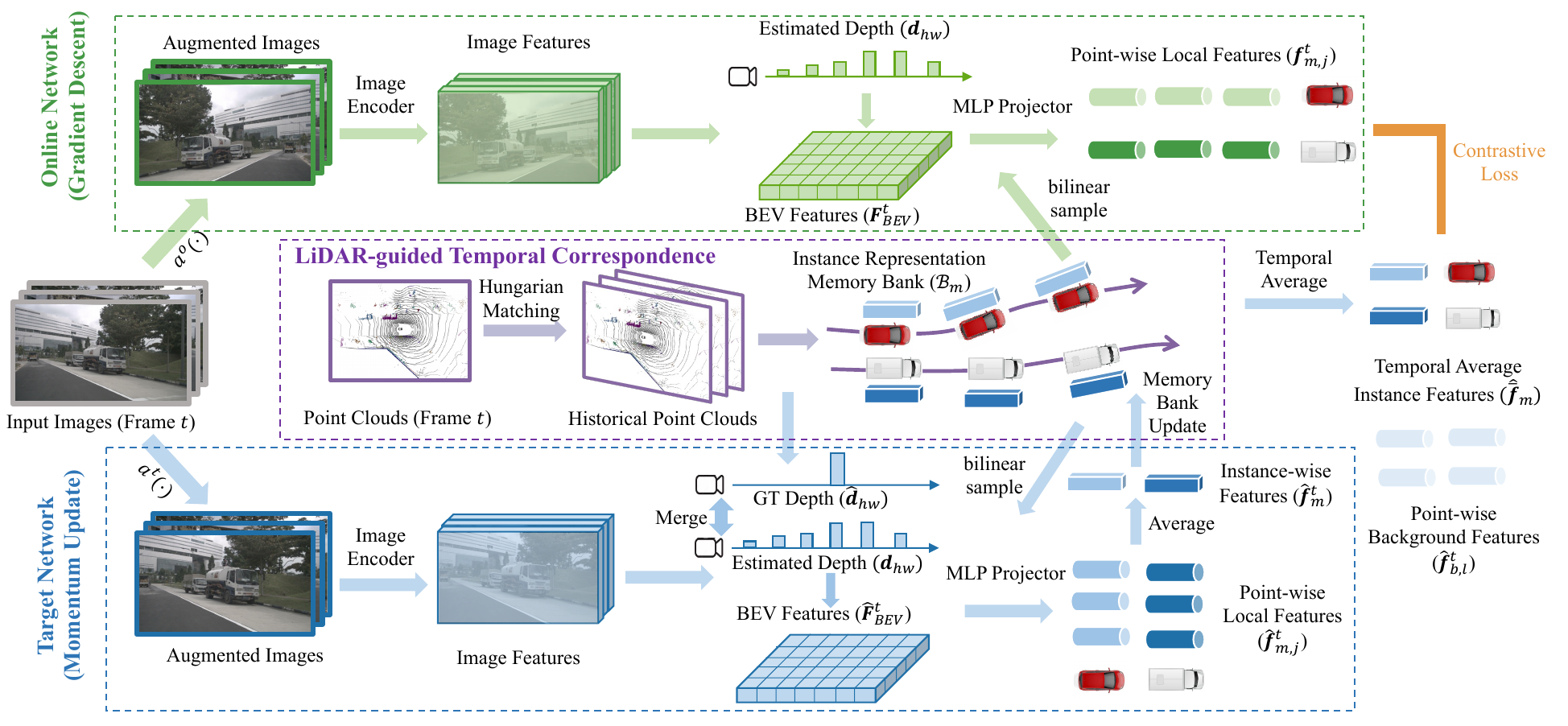}
    \caption{Overview of \ourmethod. We construct long-term instance-level temporal correspondence via unlabeled LiDAR point clouds, and design a contrastive learning framework to encourage the long-term coherence of instance representations. After pretraining, the online network is transferred to downstream task-specific models.}
    \label{fig:overview}
    \vspace{-10pt}
\end{figure*}

\paragraph{Pretraining for Perception and Prediction} Unsupervised learning algorithms~\cite{he2020momentum,grill2020bootstrap,he2022masked,bai2022point,xie2021propagate} learn expressive representation without any human annotations. Some pioneering research adapts contrastive learning~\cite{xie2020pointcontrast,huang2021spatio} or masked autoencoder~\cite{pang2022masked,zhang2022point} frameworks to 3D point clouds data, but most of them are constrained to a single-object or indoor data. To handle large outdoor scenes in autonomous driving scenarios, ProposalContrast~\cite{yin2022proposalcontrast} devises a region-level contrastive learning algorithm. BEV-MAE~\cite{lin2022bev} and GeoMAE~\cite{tian2023geomae} follow MAE~\cite{he2022masked} to predict the missed information in the BEV space. Despite the increasing popularity of cameras in autonomous driving, most methods only consider point clouds as input. Occ-BEV~\cite{min2023occ} predicts the occupancy of each voxel in the 3D space with image inputs. However, due to its lack of semantics and temporal cues, this method provides limited benefits to complex perception tasks. For prediction tasks, previous works~\cite{li2023pre,xu2022pretram} propose pretraining methods for the traditional two-stage motion prediction pipeline, which requires manually annotated historical trajectories and semantic maps as inputs.

In contrast to these works, we emphasize the long-term temporal coherence of instance-level representations and propose a novel unsupervised learning algorithm targeted for complex vision-based autonomous driving tasks.

\paragraph{Representation Learning from Temporal Cues} Previous research~\cite{wang2015unsupervised,han2020self,jabri2020space,tong2022videomae,wang2023videomae,gupta2023siamese} uses the temporal information inside videos for unsupervised representation learning. In particular, several papers share common insights with us to develop contrastive learning algorithms to utilize the fine-grained temporal correspondence. They construct the temporal correspondence using visual features~\cite{wang2015unsupervised}, optical flow~\cite{han2020self}, or cycle-consistency~\cite{jabri2020space}. However, they are limited to 2D tasks for image or video processing. TARL~\cite{nunes2023cvpr} utilizes the temporal cues for the pretraining of LiDAR-based 3D perception models, but their clustering algorithm can only retrieve short-term temporal correspondences. In contrast, our algorithm targets the long-term temporal correspondence to learn a temporally coherent instance representation for vision-based 3D tasks.

%% file: sec_arxiv/3_methods.tex
\section{Methodology}

In this section, we introduce our novel unsupervised representation learning algorithm~\ourmethod, tailored for vision-based autonomous driving tasks. Our approach utilizes unlabeled multi-view images as the primary input. Additionally, during the pretraining phase, we incorporate raw LiDAR point clouds as supplementary guidance. It is important to note that these point clouds are not required during the inference stage of the downstream tasks.

The methodology is structured into several key components: First, we lay out a straightforward formulation of our method in Sec.~\ref{sec:formulate}. Next, Sec.~\ref{sec:temporal} details how we construct temporal correspondences, which play a critical role in guiding the extraction of the bird's eye-view (BEV) features, as elaborated in Sec.~\ref{sec:feature}. Our approach is underpinned by a contrastive learning framework, which we describe in Sec.~\ref{sec:contrastive}. This framework employs an online network alongside a momentum-updated target network, aimed at learning robust representations that are temporally coherent.
An overview of~\ourmethod{} is shown in Fig.~\ref{fig:overview}.

\subsection{Formulation}
\label{sec:formulate}
Vision-based perception algorithms often resort to several past frames of input images to detect the object in 3D space. In essence, this process can be formulated as a form of structure from motion (SfM) that recovers the 3D state of an instance based on its observation from multiple perspective views at different time steps: 
\begin{equation}
    \mathbf{O}^{3D,t}_{m}=g\left(\left\{h\left(\mathbf{O}^{2D,t-k}_{m}\right)\right\}_{k=0}^K\right),
\end{equation}
where $\mathbf{O}^{3D,t}_{m}$ is the 3D state of instance $m$ at time step $t$ including its 3D size and pose, $\mathbf{O}^{2D,t}_{m}$ is the 2D visual pattern of the instance $m$ appearing on the image inputs $\mathcal{I}^t$ at frame $t$, $h(\cdot)$ encodes the 2D visual patterns at each time step separately, and $g(\cdot)$ recovers the 3D object states by combining multi-frame 2D information.

In this case, it is necessary to find the instance-wise correspondence $\left\{\mathbf{O}^{2D,t-k}_{m}\right\}_{k=0}^K$, \textit{i.e.} we need to identify the same instance $m$ across different time steps. Due to the motion of the ego-vehicle and surrounding scenes, the shape and appearance of each instance in a 2D image can vary significantly at different time steps. Given a set of instances $\left\{\mathbf{O}^{2D,t}_{m}\right\}_{m=1}^M$ at each frame $t$, it is difficult to correlate each of them with the instances at other frames across a long-term temporal sequence.

To ensure the effective fusion of multi-frame inputs, we should encourage the model $h(\cdot)$ to learn coherent instance-wise features robust to temporal changes, facilitating the instance-wise temporal correspondence in downstream tasks. At each frame $t$, the image features are projected into the BEV space as $\mathbf{F}_{BEV}^t$~\cite{philion2020lift}. We extract the instance representation $\mathbf{f}^t_m$ corresponding to each instance $\mathbf{O}^{2D,t}_{m}$ from the BEV feature map. The goal of our approach is to make the same instance representation $\{\mathbf{f}^{t-k}_m\}_{k=0}^K$ consistent in a long sequence and make the representations of different objects $\{\mathbf{f}^{t}_m\}_{m=1}^M$ distinct from each other. Concretely, we compute the temporal average representation $\bar{\mathbf{f}}_m$ for each instance $m$ over multiple historical frames as follows:
\begin{equation}
    \bar{\mathbf{f}}_m=\frac{1}{K+1}\sum_{k=0}^{K}\mathbf{f}^{t-k}_m.
    \label{eq:temporal}
\end{equation}
Our representation encourages the instance representation $\mathbf{f}^t_m$ to be near the temporal average representation $\bar{\mathbf{f}}_m$ and distinct from other instance representations $\bar{\mathbf{f}}_n,n\neq m$.

\begin{figure}
    \centering
    \includegraphics[width=\linewidth]{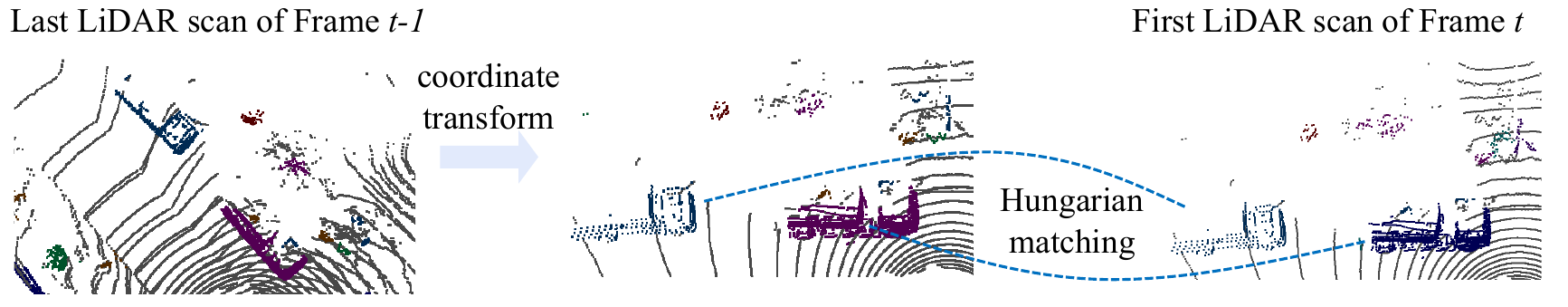}
    \caption{Construction of long-term temporal correspondence. The first and last scans corresponding to each frame are correlated by clustering. The Hungarian algorithm connects every two neighboring frames to generate long-term correspondence.}
    \label{fig:cluster}
    \vspace{-10pt}
\end{figure}

\subsection{LiDAR-guided Temporal Correspondence}
\label{sec:temporal}
Given the varying appearance of each instance at different input frames, it is difficult to construct the instance-level temporal correspondence from pure vision inputs without any explicit human annotations. Alternatively, we build the long-term temporal correspondence using raw LiDAR point clouds $\mathcal{P}^t$.
The LiDAR sensor can have a higher frequency than the cameras, so there may be multiple LiDAR scans within the small interval between two camera frames. 
For the camera image captured at time step $t$, which we refer to as frame $t$, we collect all LiDAR scans captured between $t-1$ and $t$, to create the point clouds for frame $t$, denoted as $\mathcal{P}^t$.
% For simplicity, we take the first scan captured immediately after $t-1$, $\mathcal{P}^{t,s}$, and the scan captured immediately before $t$, $\mathcal{P}^{t,e}$, and combine them together, $\mathcal{P}^t=\mathcal{P}^{t,s}\cup\mathcal{P}^{t,e}$, to create the point clouds for frame $t$.
% We denote point clouds from the first and last LiDAR scans between camera frames $t-1$ and $t$ as $\mathcal{P}^{t,s}$ and $\cup\mathcal{P}^{t,e}$ separately, which are combined as the point clouds corresponding to camera frame $t$ as $\mathcal{P}^t=\mathcal{P}^{t,s}\cup\mathcal{P}^{t,e}$.
% We correspond all the LiDAR scans between camera frames $t$ and $t-1$ to the frame $t$, only consider the first and last scans of between camera frames $t$ and $t-1$ for the sake of simplicity, denoted as $\mathcal{P}^t=\mathcal{P}^{t,s}\cup\mathcal{P}^{t,e}$.
% For the sake of simplicity, we only consider the first and last scans of between camera frames $t$ and $t-1$, which is corresponded to camera frame $t$, denoted as $\mathcal{P}^t=\mathcal{P}^{t,s}\cup\mathcal{P}^{t,e}$.
% Without any explicit human annotations, we aim to build the long-term temporal correspondence using point clouds 
We apply a two-stage strategy to retrieve the long-term temporal correspondence (Fig.~\ref{fig:cluster}). In the first stage, we identify different instances from the LiDAR scans corresponding to frame $t$. In the second stage, the inter-frame instance-wise correspondences are generated by matching the identified instances in the first stage.

\paragraph{Instance Identification}
We apply the method in~\cite{himmelsbach2010fast} to remove the ground points (black points in Fig.~\ref{fig:cluster}) from $\mathcal{P}^t$ to get the non-ground point clouds $\mathcal{\hat{P}}^t$. The non-ground point clouds $\mathcal{\hat{P}}^t$ are clustered with HDBSCAN algorithm~\cite{campello2013density} into several clusters $\mathcal{\hat{P}}^t=\bigcup_{m=1}^M\mathcal{C}^t_m$, where $\mathcal{C}^t_m$ includes points from multiple LiDAR scans between time step $t-1$ and $t$. We consider the points from the LiDAR scans immediately after time step $t-1$ and immediately before $t$, denoted as $\mathcal{C}^{t,s}_m$ and $\mathcal{C}^{t,e}_m$. We discard invalid clusters $m$, whose number of points is fewer than a threshold $\tau_n$ in either the first scan $\mathcal{C}^{t,s}_m$ or the last scan $\mathcal{C}^{t,e}_m$. Each cluster $\mathcal{C}^t_m$ is considered as a single instance. It is likely to have different center locations in the first and last LiDAR scans, denoted as $\mathbf{c}^{t,s}_m$ and $\mathbf{c}^{t,e}_m$, which indicates the intra-frame motion of this instance in the short term between the first and last LiDAR scans.

\paragraph{Long-term Matching} 
Given the instance-wise motions from $\{\mathbf{c}^{t,s}_m\}_{m=1}^M$ to $\{\mathbf{c}^{t,e}_m\}_{m=1}^M$ in a single frame $t$, we connect them to the previous frame $t-1$ by finding the correspondence between $\{\mathbf{c}^{t,s}_m\}_{m=1}^M$ at the beginning of frame $t$ and the instance centers $\{\mathbf{c}^{t-1,e}_m\}_{m=1}^M$ at the end of the frame $t-1$. Note that point clouds corresponding to different frames are in different coordinate systems due to the motion of the ego-vehicle. Given the rotation $\mathbf{R}$ and translation $\mathbf{p}$ of the ego-car between frame $t-1$ and $t$, the instance centers $\{\mathbf{c}^{t-1,e}_m\}_{m=1}^M$ should be transformed into the coordinate of frame $t$ as follows:
\begin{equation}
    \mathbf{\hat{c}}^{t-1,e}_m=\mathbf{R}^T\mathbf{c}^{t-1,e}_m-\mathbf{R}^T\mathbf{p}.
\end{equation}
The Hungarian algorithm~\cite{kuhn1955hungarian} is then used to find the correspondence between $\{\mathbf{c}^{t,s}_m\}_{m=1}^M$ and $\{\mathbf{\hat{c}}^{t-1,e}_m\}_{m=1}^M$. To handle the disappearance or appearance of instances, we pad empty instances with maximal distance to all the other instances. We set a threshold of $\tau_d=0.5m$ for the distance between the matched centers, where the valid matching should satisfy $||\mathbf{\hat{c}}^{t-1,e}_m-\mathbf{c}^{t,s}_m||_2\leq \tau_d$. This threshold can tolerate the possible object motion over a small interval ($<0.05s$) between the two scans, while avoiding the mismatch between different instances. The instance centers $\{\mathbf{\hat{c}}^{t,e}_m\}_m$ from the last LiDAR scan of frame $t$ will then be used to connect frame $t$ and $t+1$. In this way, we combine several intra-frame short-term instance trajectories to achieve long-term trajectories $\{\mathcal{C}_m^{t-k}\}_{k\leq K}$ for each instance $m$ by connecting $\mathbf{\hat{c}}^{t-k-1,e}_{m}$ and $\mathbf{c}^{t-k,s}_{m}$ frame by frame. Since different instances may have different life cycles, we set $K$ to be the maximal number of historical frames.

\subsection{Temporal Instance Feature}
\label{sec:feature}
For each frame $t$, we sample $N_F$ points uniformly on the BEV space from all the clusters $\left\{\mathcal{C}_m^t\right\}_{m=1}^M$ as $\left\{\mathcal{S}_m^t\right\}_{m=1}^M$ where $\mathcal{S}_m^t\subseteq \mathcal{C}_m^t$ and $\sum_{m=1}^M|\mathcal{S}_m^t|=N_F$. Given the vision-based BEV feature map $\mathbf{F}_{BEV}^t$, we apply bilinear sampling to extract the features $\mathbf{f}_{m,j}^t$ for each sampled point $\mathbf{s}_{m,j}^t\in\mathcal{S}_m^t$. At each frame $t$, we compute the instance-level representation by averaging the point-level features belonging to the same instance, \textit{i.e.}, 
\begin{equation}
    \mathbf{f}_m^t=\frac{1}{N_m}\sum_{j=1}^{N_m}\mathbf{f}_{m,j}^t,
    \label{eq:instance}
\end{equation}
where $N_m$ is the number of sampled points belonging to the instance $m$. For each instance, we maintain a memory bank $\mathcal{B}_m=[\mathbf{f}_m^{t-k}]_{k\leq K}$ for its instance representation at different time steps. The memory bank for each instance is created at the time step of its first appearance and will be updated at each following time step as long as this instance is matched at frame $t$.
Next, the temporal average representation $\bar{\mathbf{f}}_m$ of instance $m$ is calculated following Eq.~\ref{eq:temporal} based on the instance representation in the memory bank $\mathcal{B}_m$.

However, the foreground objects only account for a small part of the input images. To exploit the semantics from the whole image, we also sample $N_B$ background points uniformly on the BEV space not occupied by any valid clusters. Their corresponding features are extracted as $\{\mathbf{f}_{b,l}^t\}_{l=1,2,\dots,N_B}$. Due to their static nature, the background representations are more stable. Therefore, we skip the temporal averaging operation. 

\subsection{Depth-aware Contrastive Learning}
\label{sec:contrastive}
After obtaining instance-level features through LiDAR-guided temporal correspondence, we apply constrastive learning to encourage a consistent representation for the same instance. On top of the standard Siamese network framework, we leverage a few techniques to ensure the stability and robustness of the unsupervised pretraining, as detailed in this section. 

\paragraph{Architecture} We apply a Siamese network framework~\cite{grill2020bootstrap} with an online network $f(\cdot)$ and a target network $\hat{f}(\cdot)$. Both networks include an image backbone to encode the input images, a view transformation module to project the image features into the BEV space, and a BEV encoder to encode the BEV features. The online network is updated by gradient descent, while the target network uses the exponential moving average (EMA) of the online network. For the input image $\mathcal{I}^t$ for frame $t$, we generate two different augmented versions $a^o(\mathcal{I}^t)$ and $a^t(\mathcal{I}^t)$ with a set of spatial augmentations $\mathcal{A}$, $a^o,a^t\in\mathcal{A}$. The augmented versions are fed into the online and target networks separately. The online and target networks output the BEV features $\mathbf{F}^t_{BEV}$ and $\mathbf{\hat{F}}^t_{BEV}$, where
\begin{equation}
    \mathbf{F}^t_{BEV}=f(a^o(\mathcal{I}^t)),\quad \mathbf{\hat{F}}^t_{BEV}=\hat{f}(a^t(\mathcal{I}^t)).
\end{equation}

\paragraph{Depth-aware Representation Learning} 
Stable representations in the target network play an important role in contrastive learning~\cite{wang2022importance}. However, due to the depth ambiguity of 2D inputs, it is quite unreliable to transform the image features from the perspective view to the BEV space. 
% \begin{wraptable}{l}{0.47\linewidth}
%     \centering
%     \caption{SOLOFusion results with different depth estimation options.}
%     \label{tab:depth}
%     \vspace{-10pt}
%     \begin{tabular}{ccc}
%     \toprule
%         \textbf{Depth} & \textbf{NDS} & \textbf{mAP} \\
%         \midrule
%         Estimated & 49.7 & 40.6\\
%         One-hot GT & 43.9 & 33.8\\
%         Merged &\textbf{51.3}& \textbf{43.4}\\
%     \bottomrule
%     \end{tabular}
% \end{wraptable}
Many camera-based 3D perception models~\cite{philion2020lift,huang2021bevdet,park2022time} project the image features with the estimated depth distribution $\mathbf{d}_{hw}=\left[p_{hw}^1,p_{hw}^2,\dots,p_{hw}^D\right]$ for each pixel $(h,w)$ \textit{w.r.t.} a set of discrete depths defined by $[d_0+\Delta, \dots, d_0+D\Delta]$ and $\sum_{i=1}^Dp_{hw}^i=1$.
To get more reliable supervision in the pretraining stage, we combine the ground-truth depth distribution $\hat{\mathbf{d}}_{hw}$ from the LiDAR sensor with the estimated depth distribution $\mathbf{d}_{hw}$, where $\hat{\mathbf{d}}_{hw}=\left[\hat{p}_{hw}^1,\hat{p}_{hw}^2,\dots,\hat{p}_{hw}^D\right]$ is a one-hot distribution with only $\hat{p}_{hw}^k=1$ if $d_0+k\cdot\Delta$ is the ground-truth depth. We name this index to $k_{GT}$ and merge the ground-truth depth and estimated depths by setting $p_{hw}^{k_{GT}}=1$ in $\mathbf{d}_{hw}$ and then re-normalizing the merged depth distribution to ensure $\sum_{i=1}^Dp_{hw}^i=1$. 
% As shown in Tab.~\ref{tab:depth}, the merged depth brings better performance to existing vision-based detectors without any finetuning. 
We only merge the estimated and ground-truth depth for the target network while the online network only uses the estimated depth distribution. As a result, the target network can achieve much better positional precision in the BEV space, enabling a more reliable self-supervision in the contrastive learning framework.

\paragraph{View Transformation Mask}
We add a dropout mask in the view transformation process by setting a percentage, $r$, of the estimated depth distribution $\left[p_{hw}^1,p_{hw}^2,\dots,p_{hw}^D\right]$ to zero. This dropout strategy is only adopted to the online network, since the target network prefers a more stable representation for self-supervision. In this way, we encourage the network to learn stronger BEV representation to compensate for the missing information.

\paragraph{Contrastive Loss Function} The basic goal of our unsupervised learning is to encourage the instance-wise representation to be invariant to temporal changes. We extract the point-level features at the current time step $t$ from the BEV features of online network and target network respectively as in Sec.~\ref{sec:feature}. The temporal average feature $\hat{\bar{\mathbf{f}}}_m$ of each instance $m$ is derived from the target network point-level features $\hat{\mathbf{f}}_{m,j}^t$ for each sampled point $\mathbf{s}^t_{m,j}$ following Eq.~\ref{eq:instance} and Eq.~\ref{eq:temporal}. We get inspiration from~\cite{sun2023calico} to apply a point-to-instance contrastive loss as follows:
\noindent
\begin{equation}
\resizebox{0.9\linewidth}{!}{
$\mathcal{L}=\frac{1}{N_F}\underset{\mathbf{s}_{m,j}^t\in\mathcal{S}_m^t}{\sum}\frac{\exp(\mathbf{f}_{m,j}^{tT}\hat{\bar{\mathbf{f}}}_m/\tau)}{\sum_{n=1}^M\exp(\mathbf{f}_{m,j}^{tT}\hat{\bar{\mathbf{f}}}_n/\tau)+\sum_{l=1}^{N_B}\exp(\mathbf{f}_{m,j}^{tT}\hat{\mathbf{f}}_{b,l}^t/\tau)},$
}
\end{equation}
where $\tau$ is the temperature set as $0.1$ and all the features are normalized. This loss will make the local features belonging to the same instances close to each other across the long-term temporal sequence and keep them distant from other instance features as well as background features.

%% file: sec_arxiv/4_experiments.tex
\section{Experiments}
\label{sec:experiments}
This section presents a comprehensive evaluation of our proposed method. We begin by detailing the experimental setup, including dataset, and pretraining/finetuning setups in Sec.~\ref{sec:implement}. Our approach is then compared against established baselines to demonstrate its effectiveness in vision-based autonomous driving tasks. Specifically, we evaluate our method on 3D object detection (Sec.~\ref{sec:detection}), HD map construction (Sec.~\ref{sec:mapping}), motion prediction (Sec.~\ref{sec:prediction}), and planning (Sec.~\ref{sec:planning}). Additionally, we provide an analysis of our algorithm in Sec.~\ref{sec:analysis} and ablation studies in Sec.~\ref{sec:ablation}.
% In this section, we evaluate our algorithm on several camera-based perception and prediction tasks. We follow the standard protocol of unsupervised learning to first pretrain the network and then finetune it with some task models. We explain some implementation details in Sec.~\ref{sec:implement}. Then, we evaluate our method on 3D object detection (Sec.~\ref{sec:detection}), BEV semantic segmentation (Sec.~\ref{sec:segmentation}), and motion prediction (Sec.~\ref{sec:prediction}). Finally, we provide some analysis in Sec.~\ref{sec:analysis} and ablation studies in Sec.~\ref{sec:ablation}.

\subsection{Implementation Details}
\label{sec:implement}

\paragraph{Dataset} 
We use the nuScenes dataset~\cite{caesar2020nuscenes} for both pretraining and finetuning. It is the most common benchmark for vision-based autonomous driving tasks, including 700/150/150 scenes for training/validation/test. It provides point clouds collected by a 32-beam LiDAR and multi-view images from six cameras. For pretraining, we only use the raw point clouds and images from the training set without any annotations. 

\paragraph{Unsupervised Pretraining} 
In the pretraining stage, we follow the network architecture of short-term SOLOFusion~\cite{park2022time} (ResNet50 backbone~\cite{he2016deep}) without detection heads. Different from 2D representation learning~\cite{he2020momentum,grill2020bootstrap}, strong data augmentation has potential negative effects in our setting, bringing ambiguity in 3D geometry recovery from images~\cite{brazil2023omni3d}. Thus, we only apply weak data augmentations same as~\cite{park2022time} to the inputs, including random crops, scaling, and rotations of the images, as well as random rotation and scaling for the BEV representation. We sample $N_F=N_B=1000$ foreground and background points from the BEV space at each frame. The momentum of the target network is set as $0.99$. To learn the long-term representation coherence, we use at most $K=16$ (8 seconds) historical frames. A dropout percentage of $r=0.3$ is adopted in the view transformation of the online network. We pretrain the model for 24 epochs using Adam optimizer~\cite{kingma2014adam} with a batch size of 32 and a learning rate of 1e-4 on four NVIDIA A100 40GB GPUs.

\paragraph{Supervised Finetuning} We initialize the network for different tasks with our pretrained weight in the supervised finetuning stage. We follow the same training strategies as the from-scratch training for each task model.

\begin{table}[tb]
    \centering
    \caption{3D object detection using SOLOFusion~\cite{park2022time} on nuScenes}
    \label{tab:solofusion}
    \begin{subtable}[t]{\linewidth}
        \centering
        \caption{Short-term model variant}
        \label{tab:solofusion-short}
        \begin{tabular}{c|c|cc}
            \toprule
                \textbf{Finetune Percent} & \textbf{Pretraining Strategy} & \textbf{NDS} & \textbf{mAP}
                \\
                \midrule
                \multirow{2}{*}{$10\%$} & ImageNet~\cite{park2022time} & 21.3 & 15.6 \\
                & \ourmethod{} (ours) & \textbf{24.3}  & \textbf{19.4} \\
                \midrule
                \multirow{2}{*}{$20\%$} & ImageNet~\cite{park2022time} & 24.3 & 20.0\\
                & \ourmethod{} (ours) & \textbf{25.8} & \textbf{22.3}\\
                \midrule
               \multirow{2}{*}{$50\%$} & ImageNet~\cite{park2022time} & 30.8  & 28.1 \\
                & \ourmethod{} (ours) & \textbf{32.4}  & \textbf{29.1} \\
                \midrule
               \multirow{5}{*}{$100\%$} & ImageNet~\cite{park2022time} & 39.1 & 34.4 \\
               & BYOL~\cite{grill2020bootstrap} & 37.2 & 33.2\\
               & MonoDepth2~\cite{godard2019digging} & 36.4 & 32.1\\
               & Occ-BEV~\cite{min2023occ} & 39.0 & 34.3 \\
                & \ourmethod{} (ours) &  \textbf{41.3} & \textbf{35.1} \\    
                \bottomrule
        \end{tabular}
    \end{subtable}

    \begin{subtable}[t]{\linewidth}
        \centering
        \caption{Long-term model variant}
        \begin{tabular}{c|c|cc}
            \toprule
                \textbf{Finetune Percent} & \textbf{Pretraining Strategy} & \textbf{NDS} & \textbf{mAP}
                \\
                \midrule
                \multirow{2}{*}{$10\%$} & ImageNet~\cite{park2022time} & 20.6 & 13.2 \\
                & \ourmethod{} (ours) & \textbf{22.9} & \textbf{16.4}\\
                \midrule
                \multirow{2}{*}{$20\%$} & ImageNet~\cite{park2022time} & 28.5 & 23.1 \\
                & \ourmethod{} (ours) & \textbf{32.3} & \textbf{25.0} \\
                \midrule
               \multirow{2}{*}{$50\%$} & ImageNet~\cite{park2022time} & 40.9  & 32.7 \\
                & \ourmethod{} (ours) & \textbf{43.1} & \textbf{33.6}\\
                \midrule
               \multirow{5}{*}{$100\%$} & ImageNet~\cite{park2022time} & 49.7 & 40.6 \\
               & BYOL~\cite{grill2020bootstrap} & 48.9 & 39.7\\
               & MonoDepth2~\cite{godard2019digging} & 47.3 & 37.4 \\
               & Occ-BEV~\cite{min2023occ} & 49.2 &  40.4 \\
                & \ourmethod{} (ours) &  \textbf{50.2} & \textbf{40.8} \\    
                \bottomrule
        \end{tabular}
    \end{subtable}
\end{table}

\begin{table}[tb!]
    \centering
    \caption{3D object detection using BEVDet~\cite{huang2021bevdet} on nuScenes}
    \label{tab:bevdet}
    \begin{tabular}{c|c|cc}
    \toprule
    \textbf{Finetune Percent} & \textbf{Pretraining Strategy} & \textbf{NDS} & \textbf{mAP}\\
    \midrule
        \multirow{2}{*}{$10\%$} & ImageNet~\cite{huang2021bevdet} & 11.7  & 8.0 \\
        & \ourmethod{} (ours) & \textbf{16.2} & \textbf{10.8}\\
    \midrule
        \multirow{2}{*}{$100\%$} & ImageNet~\cite{huang2021bevdet} & 35.0 & 28.3 \\
        & \ourmethod{} (ours)& \textbf{36.1} & \textbf{28.6}\\
    \bottomrule
    \end{tabular}
\end{table}

\subsection{Object Detection}
\label{sec:detection}
To verify the effectiveness of our method on 3D perception tasks, we finetune a vision-based detector, SOLOFusion~\cite{park2022time}, for camera-based 3D object detection, using our pretrained weight. We consider both short-term and long-term variants of SOLOFusion model, where the former only considers one neighbor frame but the latter resorts to the past 16 frames. More specifically, we finetune the model using $10\%,20\%,50\%,100\%$ of the training set separately, and report the mAP and NDS metrics~\cite{caesar2020nuscenes} on the validation set. The results are reported in Tab.~\ref{tab:solofusion}.

The most important advantage of unsupervised learning is to improve data efficiency with limited annotation by utilizing other unlabeled data. Not surprisingly, for both short-term and long-term variants,~\ourmethod{} significantly outperforms the from-scratch training baseline using ImageNet weights, especially when only a subset of the training data is used for finetuning. In addition, the results show that the pretraining can provide performance gains even when finetuning the models on the entire training set. In this case, the pretraining does not include extra information but utilizes the inherent temporal cues inside the training data. Our pretraining also outperforms several other unsupervised learning baselines, including BYOL~\cite{grill2020bootstrap}, a 2D contrastive learning algorithm that does not take the 3D geometry into account, and MonoDepth2~\cite{godard2019digging} and Occ-BEV~\cite{min2023occ} that consider geometric information but ignore semantics. 
% Consequently, all these pretraining algorithms have inferior performance in comparison with ours.

The representation learned in the pretraining stage is agnostic to different network architectures. For example, we transfer our pretrained backbone to another 3D object detection model, BEVDet~\cite{huang2021bevdet}. Tab.~\ref{tab:bevdet} shows significant performance improvement brought by our pretraining when we finetune the model on both a subset and the full training set.

\subsection{HD Map Construction}
\label{sec:mapping}
Apart from detecting foreground objects, we examine our learned representation on background detection through an HD map construction task. We apply MapTR~\cite{liao2022maptr} as the task model, and finetune it on the full training set using our pretrained weights. Our pretraining improves the mAP metric from $50.3$ to $51.2$ in comparison to ImageNet pretraining. This improvement indicates that~\ourmethod{} can also help the model to better understand the background scene such as map lanes.

\begin{table}[tb!]
    \centering
    \caption{Motion prediction using VAD-tiny~\cite{jiang2023vad} on nuScenes}
    \label{tab:prediction}
    \begin{tabular}{c|cc|cc}
    \toprule
     \multirow{3}{*}{\textbf{\makecell{Pretraining\\Strategy}}} & \multicolumn{2}{c|}{\textbf{Perception}} & \multicolumn{2}{c}{\textbf{Prediction}}\\
       & \textbf{NDS} & \textbf{mAP} & \makecell{\textbf{EPA}\\(car)} & \makecell{\textbf{EPA}\\(pedestrian)}\\
    \midrule
    ImageNet~\cite{jiang2023vad} & 39.0 & 27.0 & 59.8 & 29.0 \\
    \ourmethod{} (ours)& \textbf{41.4} & \textbf{27.9} & \textbf{63.1} & \textbf{33.1}\\
    \bottomrule
    \end{tabular}
    
\end{table}

\subsection{Motion Prediction}
\label{sec:prediction}
Learning a coherent instance representation over time not only helps to determine the current 3D states of each instance, but also leads to stronger temporal reasoning, such as predicting the future motion of these instances.
To verify this hypothesis, we apply a state-of-the-art vision-based prediction and planning model, VAD~\cite{jiang2023vad}, as the task model for camera-based motion prediction task. We transfer our pretrained backbone weights as its initialization. We finetune the VAD-tiny model on nuScenes using the full train set. This model outputs both perception and end-to-end motion prediction results. The perception performance is measured by mAP and NDS, while the prediction performance is evaluated using the End-to-end Prediction Accuracy (EPA)~\cite{gu2023vip3d} metric. The results are shown in Tab.~\ref{tab:prediction}. On the perception metrics, our pretraining leads to performance gain compared to the baseline, which is consistent with our observations in Sec.~\ref{sec:detection}. Also, the end-to-end motion prediction benefits from our pretraining method, which encourages instance-level coherence over long-term historical frames. As a result, without using any extra data,~\ourmethod{} greatly lifts the performance of the vision-based end-to-end motion prediction task.

\begin{table}[tb!]
    \centering
    \caption{End-to-end Planning using VAD-tiny~\cite{jiang2023vad} on nuScenes}
    \label{tab:planning}
    \resizebox{\linewidth}{!}{
    \begin{tabular}{c|ccc|ccc}
    \toprule
      \multirow{2}{*}{\textbf{\makecell{Pretraining\\Strategy}}} & \multicolumn{3}{c|}{$L_2$ (m)} & \multicolumn{3}{c}{Collision (\%)}\\
     & $1s$ & $2s$ & $3s$ & $1s$ & $2s$ & $3s$\\
    \midrule
    ImageNet~\cite{jiang2023vad} & 0.46 & 0.76 & 1.12 & 0.21 & 0.35 & 0.58\\
    \ourmethod{} (ours)& \textbf{0.28} & \textbf{0.52} & \textbf{0.84} & \textbf{0.11} & \textbf{0.26} & \textbf{0.56}\\
    \bottomrule
    \end{tabular}
    }
\end{table}

\subsection{End-to-end Planning}
\label{sec:planning}
Our pretraining benefits the downstream task models in predicting the current (Sec.~\ref{sec:detection}) and future (Sec.~\ref{sec:prediction}) 3D states of surrounding environments. This improved scene understanding also facilitates the planning of the ego-vehicle. We apply VAD-tiny~\cite{jiang2023vad} as the baseline for the end-to-end planning task. The performance is measured by the $L_2$ error and collision rates of the future trajectory. Results in Tab.~\ref{tab:planning} demonstrate that~\ourmethod{} significantly boosts the end-to-end planning performance, through better estimates of the current and future states of surrounding agents.

\begin{figure}[tb]
    \centering
    \includegraphics[width=\linewidth]{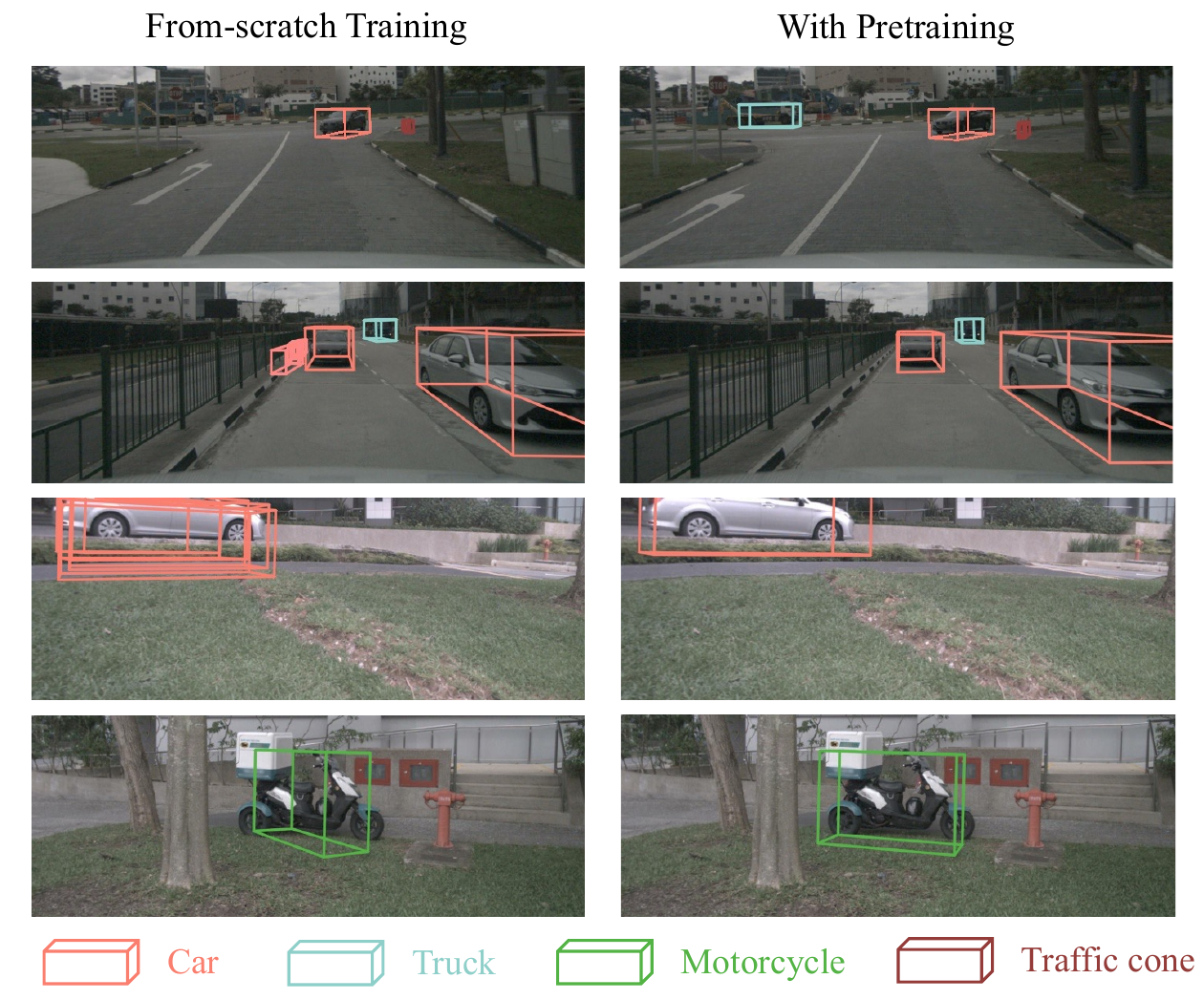}
    \caption{Qualitative comparison of SOLOFusion model trained from scratch (left) vs with pretraining  (right). 
    Our pretraining is effective at improving recall rates, reducing false positives, and localizing objects in the 3D space.
    }
    \label{fig:qualitative}
    \vspace{-10pt}
\end{figure}

\subsection{Analysis}
\label{sec:analysis}
\noindent\textbf{Qualitative Results} In Fig.~\ref{fig:qualitative}, we show some qualitative 3D object detection results using SOLOFusion~\cite{park2022time} model. Our pretraining leads to some notable improvements. First, our pretraining provides instance representation robust to the observation perspective views. This helps the detector to recall more objects. Second, our contrastive learning framework samples both foreground and background features to help learn to distinguish between the objects and background clutter. This reduces the false positive results in the downstream detection task. Finally, our temporally coherent representation benefits the multi-frame temporal fusion to achieve better 3D object pose estimation.

\begin{wrapfigure}{l}{0.6\linewidth}
    \centering
    \vspace{-15pt}
    \includegraphics[width=\linewidth]{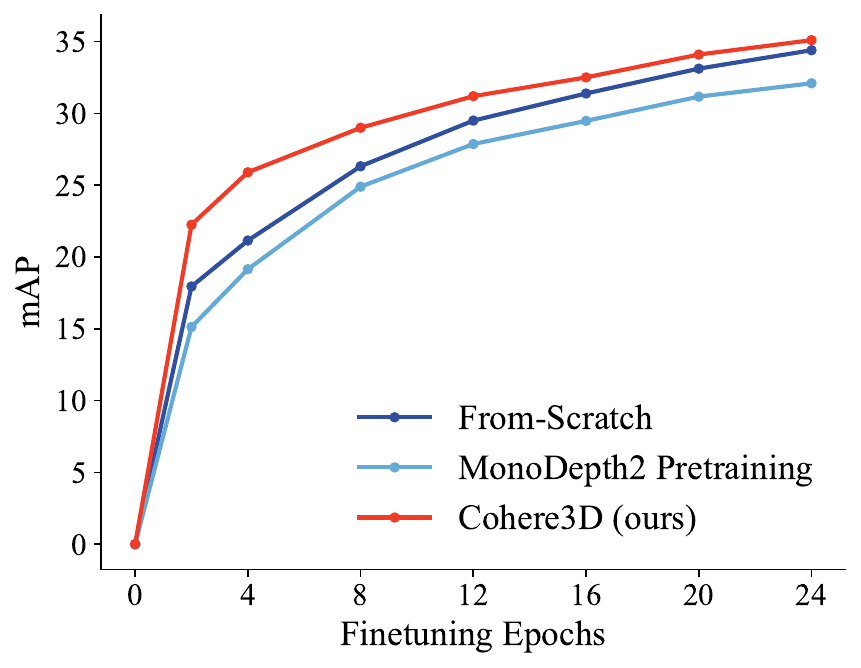}
    \caption{Performance at different finetuning epochs. Our pretraining leads to faster convergence.}
    \label{fig:speed}
    \vspace{-5pt}
\end{wrapfigure}

\noindent\textbf{Convergence Speed} Fig.~\ref{fig:speed} shows the performance of short-term SOLOFusion~\cite{park2022time} model at different training epochs during finetuning. Our pretraining leads to superior performance in the early finetuning stage than the from-scratch training. Even though we do not use any labels during pretraining, our unsupervised learning algorithm indeed learns meaningful feature representation directly transferable to the downstream task.

\begin{figure}[tb!]
    \centering
    \includegraphics[width=0.95\linewidth]{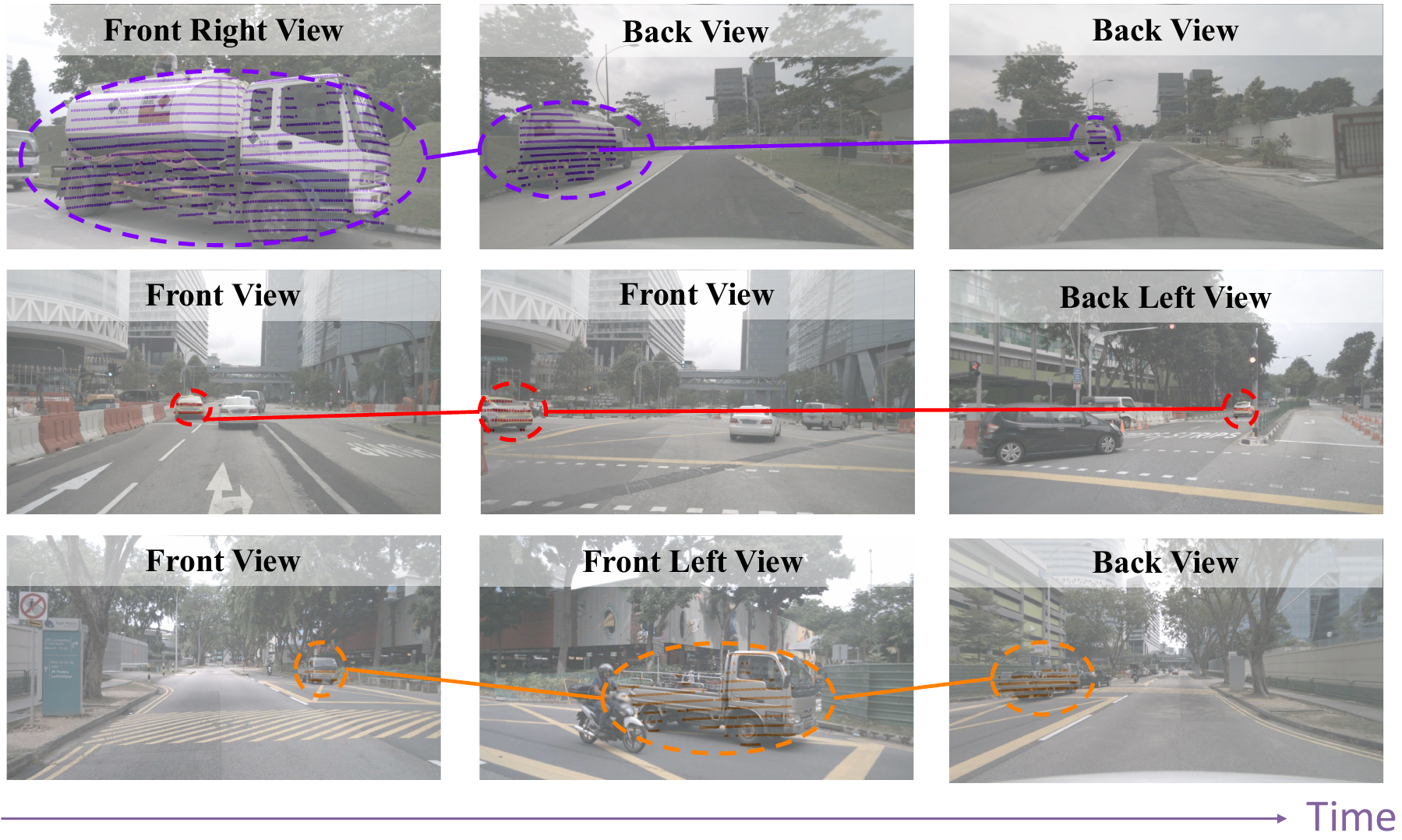}
    \caption{LiDAR-guided long-term correspondence. It can reliably track each instance over long-term historical frames even if it is captured by different cameras from varying angles. Readers may zoom in to see the projected point clouds clearly.}
    \label{fig:correspondence}
    \vspace{-10pt}
\end{figure}

\noindent\textbf{LiDAR-Guided Long-term Correspondence} Our pretraining algorithm emphasizes the long-term coherence of instance representation. This representation quality heavily depends on the accuracy of our long-term correspondence constructed in the pretraining stage. 
% Otherwise, those mismatched instances would generate much noise in the temporal average instance representation (Eq.~\ref{eq:temporal}). 
Fig.~\ref{fig:correspondence} provides some examples of long-term correspondence guided by the LiDAR point clouds. We find that our method can retrieve reliable instance trajectories without any explicit human annotations over multiple historical frames for those dynamic instances. This provides a solid foundation for our contrastive learning over temporal cues.

% \begin{table}[tb!]
%     \centering
%     \caption{Ablation study of long-term temporal coherence}
%     \label{tab:temporal}
%     \begin{tabular}{c|cc|cc}
%     \toprule
%     \multirow{2}{*}{\textbf{\makecell{Frame\\Number}}} & \multicolumn{2}{c|}{$10\%$}  & \multicolumn{2}{c}{$100\%$}\\
%     & \textbf{NDS} & \textbf{mAP} & \textbf{NDS} & \textbf{mAP} \\
%     \midrule
%     0 & 23.6 & 18.6 & 39.9 & 34.8\\
%     4 & \textbf{24.3} & 19.0 & 41.1 & 34.9 \\
%     16 & \textbf{24.3} & \textbf{19.4} & \textbf{41.3} & \textbf{35.1}\\
%     \bottomrule
%     \end{tabular}
% \end{table}

\subsection{Ablation Studies}
\label{sec:ablation}
In this part, we justify our design by examining the role of each module in our pretraining framework. We first highlight the contribution of our key component, long-term temporal coherence over instance representations. Then, we discuss the role of each design decision in the contrastive learning framework. All the experiments in this section use the short-term version SOLOFusion~\cite{park2022time} on 3D object detection task. The model is finetuned on the full nuScenes training set.

\paragraph{Long-term Temporal Coherence} 
\begin{wraptable}{r}{0.5\linewidth}
    \centering
    \caption{Ablation study of long-term temporal coherence}
    \label{tab:temporal}
    \vspace{-5pt}
    \begin{tabular}{c|cc}
    \toprule
    \textbf{\makecell{Historical\\Frame}} & \textbf{NDS} & \textbf{mAP} \\
    \midrule
    0 & 39.9 & 34.8\\
    1 & 40.1 & 34.6 \\
    4 & 41.1 & 34.9 \\
    16 &  \textbf{41.3} & \textbf{35.1}\\
    \bottomrule
    \end{tabular}
    \vspace{-5pt}
\end{wraptable}
Our algorithm concentrates on the extraction and learning of long-term instance-wise temporal correspondence. 
In the pretraining stage, we construct a memory bank for the historical instance features to encourage the temporal coherence of instance representations. We tried different numbers of historical frames in Tab.~\ref{tab:temporal}. Without any temporal information, a significant performance drop is observed. 
This verifies that long-term temporal coherence is crucial for the downstream task. Results also show that more historical frames can lead to stronger representation and bring better performance to the downstream task.

\paragraph{Contrastive Learning Modules}
We also examine the role of other modules in our contrastive learning framework, as summarized in Tab.~\ref{tab:modules}. The results show that it is important to merge the ground-truth depth with the estimated depth in the view transformation module of the target network (Sec.~\ref{sec:contrastive}). Since the depth estimation from image inputs is fragile, the ground-truth depth from the LiDAR point clouds can notably reduce the positional ambiguity in the target network's BEV features. As a result, this leads to a more stable target in the contrastive learning framework. Secondly, the extraction of background point-level features (Sec.~\ref{sec:feature}) makes full use of semantics in the input images outside of the foreground objects. It also helps to distinguish the foreground from the background. Finally, extra dropout in the LSS module~\cite{philion2020lift} for view transformation (Sec.~\ref{sec:contrastive}) results in stronger representation. Therefore, all the above design decisions contribute meaningfully to the final performance of our algorithm.

\begin{table}[tb]
    \caption{Ablation study of contrastive learning modules}
    \label{tab:modules}
    \centering
    \begin{tabular}{ccccc}
    \toprule
       \textbf{\makecell{Background\\Points}} & \textbf{\makecell{Merged\\Depth}} & \textbf{\makecell{LSS\\ Mask}} & \textbf{NDS} & \textbf{mAP} \\
    \midrule
     & \ding{51} & \ding{51} & 39.8 & 35.0\\
    \ding{51} &  & \ding{51} & 38.3 & 34.8\\
    \ding{51} & \ding{51} &  & 40.2 & 34.7\\
    \ding{51} & \ding{51} & \ding{51} & \textbf{41.3} & \textbf{35.1}\\
    \bottomrule
    \end{tabular}
\end{table}

%% file: sec_arxiv/5_conclusion.tex
\section{Conclusions}
We propose a novel contrastive learning algorithm,~\ourmethod{}, for unsupervised representation learning of vision-based 3D autonomous driving tasks. The algorithm encourages the long-term coherence of instance features across multiple frames, effectively and robustly handling changes in observation viewpoints. By utilizing unlabeled LiDAR point clouds to establish long-term instance-level correspondences,~\ourmethod{} enhances the feature sampling from the BEV space. Through comprehensive experiments, we show that our algorithm not only improves data efficiency, but also significantly boosts the final performance of downstream 3D perception, prediction, and planning tasks.

%% file: sec_arxiv/X_suppl.tex
\clearpage
\setcounter{page}{1}
\maketitlesupplementary

In this supplementary material, we elaborate on the details of our pretraining and finetuning in Sec.~\ref{sec:extra_details}. We then visualize additional qualitative results for end-to-end planning in Sec.~\ref{sec:extra_visualize}. Finally, the limitations and future work are discussed in Sec.~\ref{sec:limitation}.

\section{Implementation Details}
\label{sec:extra_details}
In this part, we present extra implementation details in addition to Sec.~\ref{sec:implement}.

\subsection{Pretraining Architecture}
In the pretraining stage, we follow the network architecture of SOLOFusion~\cite{park2022time} without the detection heads and the long-term BEV feature concatenation component. The multi-view camera inputs are encoded by an image backbone separately. We then perform a stereo matching of the encoded features with the image features from the previous frame, to construct a cost volume to improve the per-frame depth estimation. Using the image features and the depth estimation results, we generate a BEV feature map through LSS-style~\cite{philion2020lift} projection. The output is encoded by a lightweight convolutional neural network to derive the BEV features $\mathbf{F}^t_{BEV}$. We follow prior contrastive learning works~\cite{grill2020bootstrap,caron2021emerging} to design a projection head and a prediction head for the point-wise features sampled from the BEV features. Both heads are two fully-connected layers with a LayerNorm layer and a ReLU layer in between. The projection head is attached to both online and target networks, while the prediction head is only included in the online network, following the projection head.

\subsection{Pretraining Data Augmentation}
Same as~\cite{huang2021bevdet,park2022time}, we perform weak data augmentations on both image inputs and the projected BEV features. For the image inputs, we apply random resizing $(0.94,1.11)$, random rotation $(-5.4^{\circ},5.4^{\circ})$, and random horizontal flipping. For the BEV features, we apply random resizing $(0.95,1.05)$, random rotation $(-22.5^{\circ},22.5^{\circ})$, and random horizontal/vertical flipping.

\subsection{Finetuning Details}
In this part, we present the training strategies used during the finetuning stage of each task, which are the same as their default from-scratch training settings.

\paragraph{Object Detection} For both SOLOFusion~\cite{park2022time} and BEVDet~\cite{huang2021bevdet} models, they are finetuned for 24 epochs using a learning rate of 2e-4, a batch size of 64, and the AdamW optimizer~\cite{loshchilov2018fixing}. 

\paragraph{Map Reconstruction} 
The MapTR~\cite{liao2022maptr} model is finetuned for 24 epochs using a batch size of 32 and the AdamW optimizer~\cite{loshchilov2018fixing}. The learning rate is 6e-4 for the entire model except 6e-5 for the image backbone, and cosine annealing is applied to the learning rate.

\paragraph{Prediction and Planning}
We use the VAD-tiny~\cite{jiang2023vad} model for both motion prediction and end-to-end planning. The model is finetuned for 60 epochs using a batch size of 8 and the AdamW optimizer~\cite{loshchilov2018fixing}. The learning rate is 2e-4 for the entire model except 2e-5 for the image backbone, and cosine annealing is applied to the learning rate.

\section{E2E Planning Qualitative Results}
\label{sec:extra_visualize}
\begin{figure*}[tb]
    \centering
    \begin{subfigure}{0.9\linewidth}
        \centering
        \includegraphics[width=\linewidth]{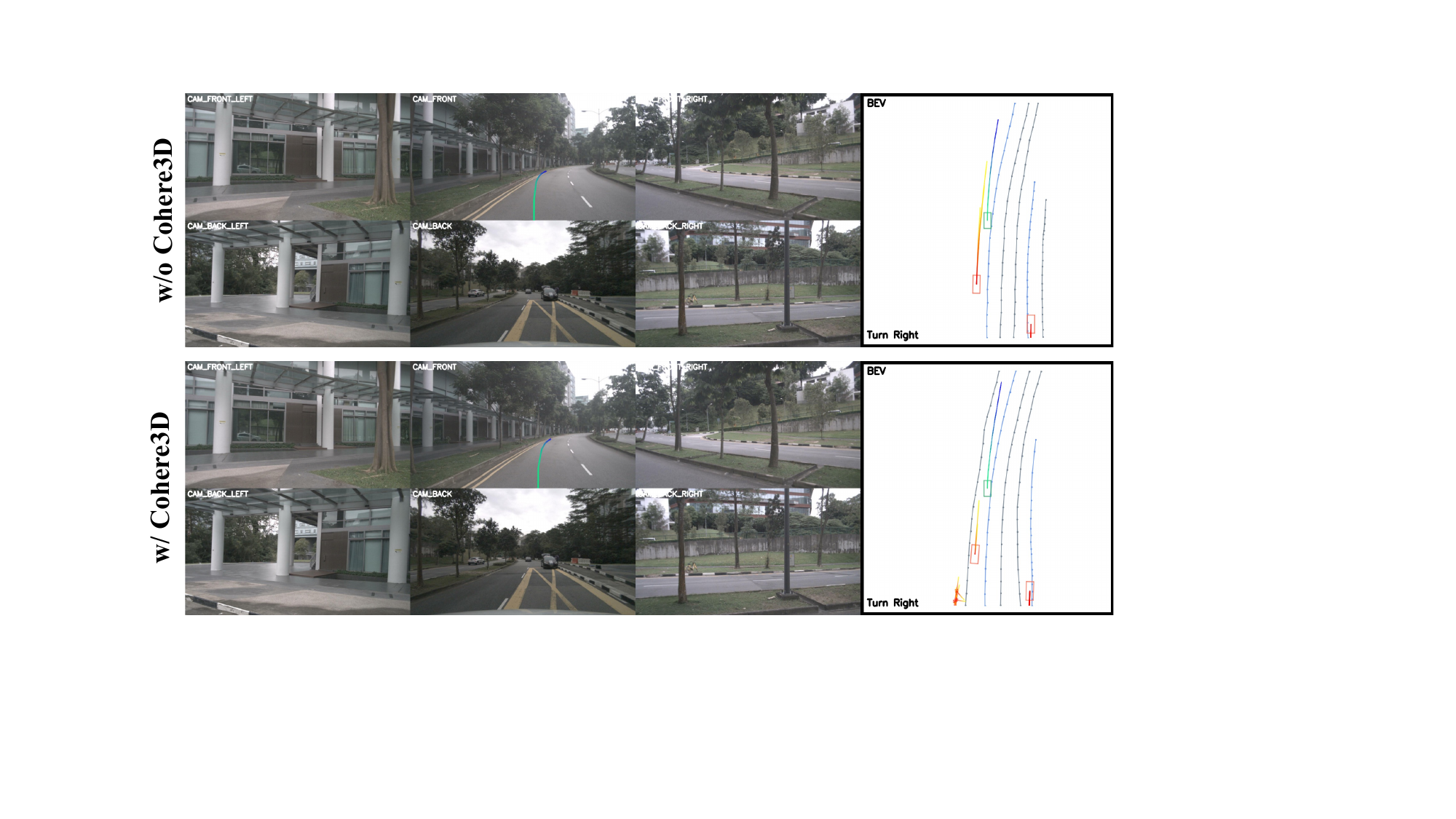}
        \caption{Cohere3D pretraining identifies the boundary of the road to provide a map-compliant plan for turning.}
    \end{subfigure}
    \begin{subfigure}{0.9\linewidth}
        \centering
        \includegraphics[width=\linewidth]{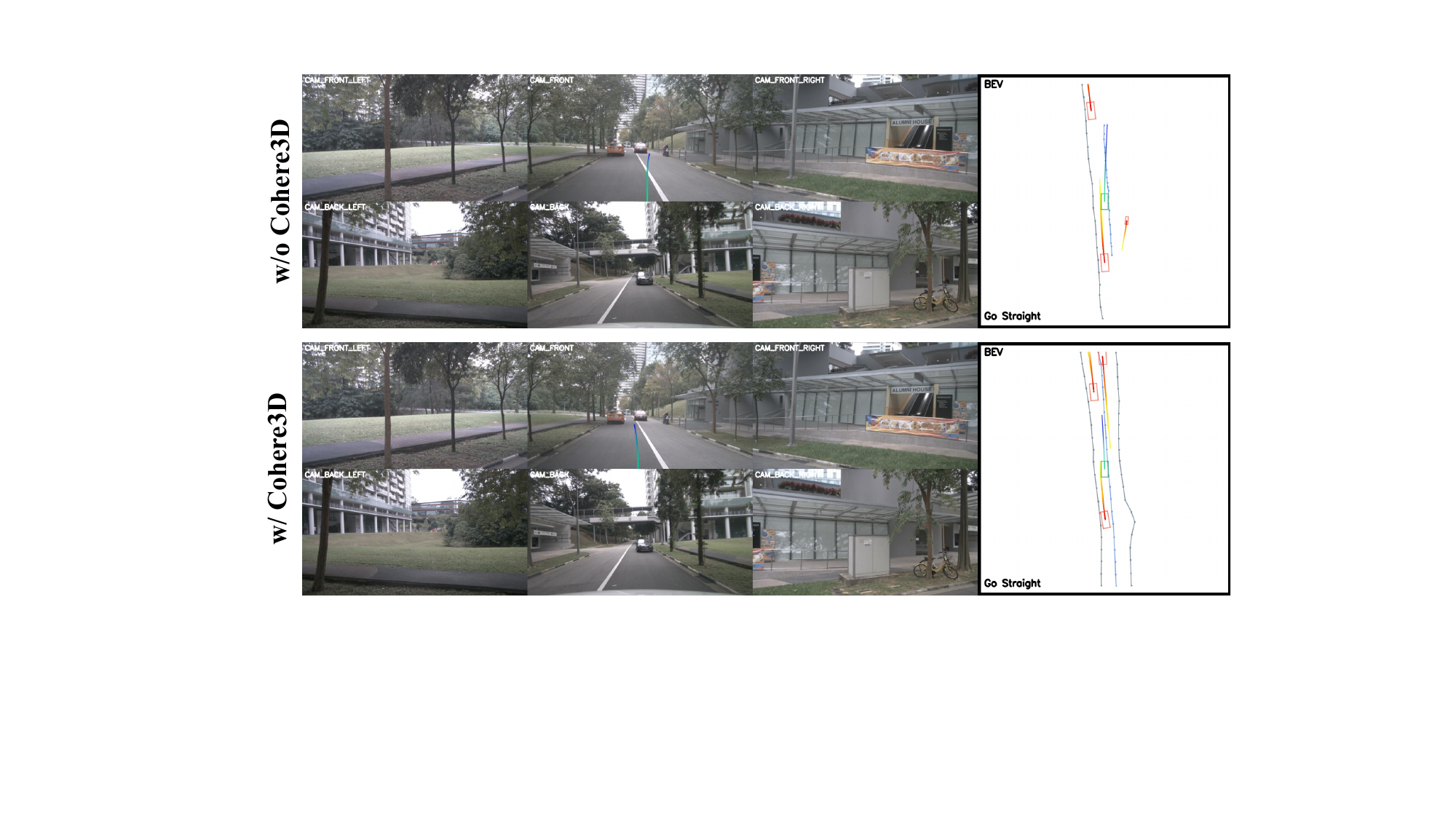}
        \caption{Cohere3D pretraining detects the car from the opposite direction to avoid potential conflicts during planning.}
    \end{subfigure}

    \caption{Visualizations of end-to-end planning results without (top) and with (bottom) Cohere3D pretraining. \textcolor{green}{Green} denotes the ego-vehicle and planning results. \textcolor{red}{Red} refers to the surrounding vehicles and prediction results. Black lines and \textcolor{blue}{blue} lines are the predicted road boundaries and predicted lane dividers, respectively.}
    \label{fig:vis_pred}
    \vspace{50pt}

\end{figure*}

In this part, we provide two qualitative comparisons of the VAD-tiny~\cite{jiang2023vad} model for end-to-end planning with and without our Cohere3D pretraining in Fig.~\ref{fig:vis_pred}. These end-to-end planning examples reflect the effectiveness of our Cohere3D pretraining for all tasks, including detection, mapping, prediction, and planning. Results show that our Cohere3D pretraining improves the performance of the end-to-end planning model in different aspects, including (1) reconstructing accurate HD maps from the camera inputs, (2) detecting surrounding vehicles and predicting their future motions reliably, and (3) producing map-compliant and safe planning results in different scenarios. 

\section{Limitation and Future Work}
\label{sec:limitation}
Although Cohere3D pretraining demonstrates promising results in multiple tasks, there are still several limitations. Firstly, our pretraining relies on pairwise raw LiDAR and multi-view camera sensors, since we require point clouds to construct the long-term fine-grained correspondence. Secondly, our experiments focus primarily on nuScenes dataset~\cite{caesar2020nuscenes} due to availability of high-quality LiDAR-image pairwise data in open-source driving datasets.

Our future work should focus on solving these limitations. For instance, we can potentially leverage unsupervised tracking models~\cite{wang2019unsupervised,jabri2020space} to replace the LiDAR point clouds for the construction of long-term correspondence. Additionally, we will explore the transferability of Cohere3D on more driving data in the future.